\documentclass{article}

\usepackage{microtype}
\usepackage{graphicx}
\usepackage{subcaption}
\usepackage{booktabs} 
\usepackage{multirow} 
\usepackage{amsfonts}       
\usepackage{nicefrac}       

\usepackage{hyperref}
\usepackage{tabularx}

\usepackage[preprint]{icml2026}

\usepackage{amsmath}
\usepackage{amssymb}
\usepackage{mathtools}
\usepackage{amsthm}

\usepackage{colortbl} 
\definecolor{GrayLine}{gray}{0.9} 
\usepackage{enumitem}
\usepackage[capitalize,noabbrev]{cleveref}

\theoremstyle{plain}

\theoremstyle{definition}

\theoremstyle{remark}

\usepackage[textsize=tiny]{todonotes}

\icmltitlerunning{Universal Anti-forensics Attack against Image Forgery Detection via Multi-modal Guidance}

\begin{document}

\twocolumn[
\icmltitle{Universal Anti-forensics Attack against Image Forgery Detection\\ via Multi-modal Guidance}



\icmlsetsymbol{equal}{*}

\begin{icmlauthorlist}
\icmlauthor{Haipeng Li}{equal,szu}      
\icmlauthor{Rongxuan Peng}{equal,szu,ntu} 
\icmlauthor{Anwei Luo}{ntu}         
\icmlauthor{Shunquan Tan}{msu_bit}          
\icmlauthor{Changsheng Chen}{msu_bit}          
\icmlauthor{Anastasia Antsiferova}{msu}          
\end{icmlauthorlist}

\icmlaffiliation{szu}{Shenzhen University, China}
\icmlaffiliation{ntu}{Nanyang Technological University, Singapore}
\icmlaffiliation{msu_bit}{Shenzhen MSU-BIT University, China}
\icmlaffiliation{msu}{Lomonosov Moscow State University's Institute for Artificial Intelligence, Russia}

\icmlcorrespondingauthor{Shunquan Tan}{tansq@smbu.edu.cn}

\icmlkeywords{Anti-Forensics, AIGC Detection, Vision-Language Models, Adversarial Attack, Explainable Forensics, Foundation Model Securityn}

\vskip 0.3in
]


\printAffiliationsAndNotice{\icmlEqualContribution}

\begin{abstract}
The rapid advancement of AI-Generated Content (AIGC) technologies poses significant challenges for authenticity assessment. However, existing evaluation protocols largely overlook anti-forensics attack, failing to ensure the comprehensive robustness of state-of-the-art AIGC detectors in real-world applications. To bridge this gap, we propose ForgeryEraser, a framework designed to execute universal anti-forensics attack without access to the target AIGC detectors. We reveal an adversarial vulnerability stemming from the systemic reliance on Vision-Language Models (VLMs) as shared backbones (e.g., CLIP), where downstream AIGC detectors inherit the feature space of these publicly accessible models. Instead of traditional logit-based optimization, we design a multi-modal guidance loss to drive forged image embeddings within the VLM feature space toward text-derived authentic anchors to erase forgery traces, while repelling them from forgery anchors. Extensive experiments demonstrate that ForgeryEraser causes substantial performance degradation to advanced AIGC detectors on both global synthesis and local editing benchmarks. Moreover, ForgeryEraser induces explainable forensic models to generate explanations consistent with authentic images for forged images. Our code will be made publicly available.
\end{abstract}

\begin{figure}[t]
  \centering
  \includegraphics[width=\columnwidth]{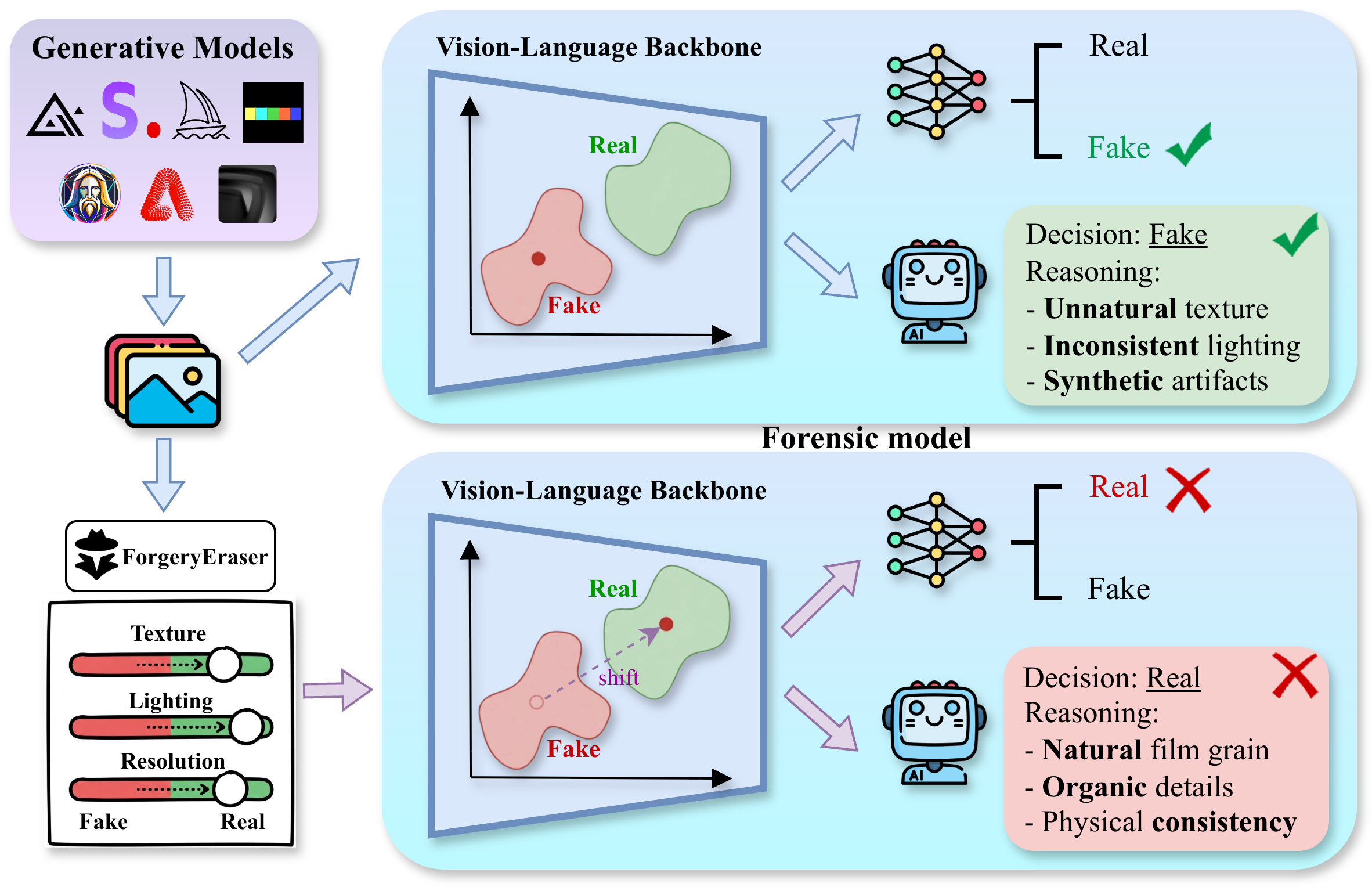}
  
  \caption{\textbf{Universal Anti-forensics Attack with ForgeryEraser.}
  \textbf{Top:} A standard forensic model correctly identifies synthetic artifacts. 
  \textbf{Bottom:} By guiding embeddings within the shared backbone toward authentic anchors, our method causes downstream detectors to invert their verdicts and fabricate plausible justifications.}
  \label{fig:teaser}
\end{figure}

\section{Introduction}

The advancement of generative AI, particularly diffusion models and Generative Adversarial Networks (GANs), has significantly lowered the barrier for creating high-fidelity AI-Generated Content (AIGC)~\cite{Karras2020styleganCVPR, Rombach2022ldmCVPR,Esser2024sd3ICML}. As synthesis algorithms iterate rapidly, the diversity and realism of generated imagery continue to expand, increasingly obscuring the boundary between reality and fabrication~\cite{Mirsky2021surveyACM}.

To address the potential misuse of these evolving generators, the forensic community has focused on improving generalization as a critical objective~\cite{Ojha2023universalCVPR}. Conventional supervised paradigms, typically trained from scratch, rely heavily on detecting low-level statistical anomalies~\cite{Rossler2019faceppICCV,Wang2020cnnCVPR}. However, restricted by homogeneous training data, these models often overfit to specific forgery patterns within a low-ranked feature space, causing significant performance degradation on unseen architectures~\cite{geirhos2020shortcut, Yan2025effortICML}. To address this, recent approaches incorporate pre-trained foundation models (particularly VLMs like CLIP~\cite{Radford2021clipICML}) as backbones specifically to leverage their higher-ranked representations and robust semantic priors for AIGC detection~\cite{Yan2025effortICML}. Consequently, these advanced detectors can identify high-level inconsistencies, such as violations of physical laws, that are often imperceptible to low-level statistical tools~\cite{xu2025fakeshieldICLR}. Furthermore, this semantic depth supports the development of interpretable forensics. By integrating VLMs with Large Language Models (LLMs), recent works can provide detailed textual reasoning to support their detection verdicts~\cite{Wen2025fakevlmNIPS, Huang2025sidaCVPR}.

However, despite these improvements in generalization, the robustness of current protocols against anti-forensics attack remains largely unexplored~\cite{Peng2025forensicsSAM}. In this paper, we reveal a systemic vulnerability arising from the widespread reliance on shared upstream backbones (e.g., CLIP). Because diverse downstream detectors integrate these publicly accessible encoders, they inherit the semantic feature space of the upstream model. This dependency fundamentally alters the threat landscape: adversarial perturbations optimized solely on the accessible upstream backbone can effectively transfer to downstream tasks. Consequently, instead of training specific surrogate models, we can execute universal anti-forensics attack by directly manipulating these inherited representations, without requiring access to the parameters of diverse downstream detectors designed for specific forensic tasks.

Crucially, current anti-forensics research has largely overlooked this specific vulnerability. Traditional methods are primarily optimized to suppress low-level statistical artifacts in DeepFake scenarios. However, they exhibit limited transferability to the broader AIGC domain due to the fundamental divergence in feature representations~\cite{huang2020fakepolisherACM, jia2022frequencyCVPR}. On the other hand, existing adversarial attacks against VLMs typically prioritize altering semantic content (e.g., object labels)~\cite{hu2024prm, fang2024cpgcICCV}, creating a fundamental task discrepancy that hinders their transferability to detectors tasked with distinguishing real from fake images.

To bridge this gap, we propose ForgeryEraser (Figure~\ref{fig:teaser}), a framework designed to execute universal anti-forensics attack against AIGC detectors. To verify the impact of this vulnerability, we select the CLIP model as our upstream backbone, driven by its widespread use in modern forensics. Leveraging the multi-modal capabilities of this backbone, we design a multi-modal guidance loss to guide the optimization of adversarial perturbations. Specifically, we define sets of text prompts describing authentic attributes and forgery attributes, which are encoded by the text encoder of CLIP into semantic anchors. The goal is to pull the forged embeddings closer to the authentic anchors while pushing them away from the forgery anchors, effectively erasing forgery traces within the CLIP feature space. Furthermore, leveraging the prior knowledge of the generative type, we employ a source-aware strategy to tailor the optimization: for global synthesis, we repel features from global synthesis attributes, whereas for local editing, we target local forgery attributes. Consequently, ForgeryEraser achieves universal anti-forensics attack against diverse AIGC detectors without requiring access to their specific parameters.

Our main contributions are summarized as follows:
\begin{itemize}[topsep=4pt, itemsep=4pt, partopsep=2pt, parsep=3pt]
\item We identify that the widespread reliance on shared upstream backbones (e.g., CLIP) creates a systemic vulnerability for AIGC detectors. This dependency enables adversaries to execute universal anti-forensics attack via a direct upstream surrogate, allowing for transferable attacks without accessing the downstream detectors.

\item We propose ForgeryEraser, a universal framework driven by the multi-modal guidance loss. Using a source-aware strategy, our method effectively erases forgery traces in images generated by both Global Synthesis and Local Editing within the CLIP feature space.

\item Extensive experiments demonstrate that ForgeryEraser causes substantial performance degradation to advanced AIGC detectors. Moreover, it induces explainable forensic models to fabricate authentic justifications for forged images.
\end{itemize}

\section{Related Work}

\subsection{Deep Learning-based Image Forensics}
\textbf{Conventional Supervised Forensics.} Prior to the foundation model era, forensic paradigms relied on training specialized detectors from scratch using supervised learning. These methods aimed to capture intrinsic statistical anomalies directly from pixel-level data. For instance, classifiers were trained to identify specific artifacts in GAN-generated imagery~\cite{Wang2020cnnCVPR} or facial manipulation traces in Deepfakes~\cite{afchar2018mesonetWIFS,Rossler2019faceppICCV}. Similarly, early IFDL networks pinpointed regional inconsistencies, such as splicing boundaries, by learning solely from annotated tampering masks~\cite{Guillaro2023truforCVPR, Peng2024codeTIFS}. However, lacking external knowledge, these models typically learn low-ranked feature representations that overfit to specific forgery patterns and suffer from limited robustness against unseen generative architectures~\cite{Frank2020frequencyICML, Yan2025effortICML}.

\textbf{Foundation Model-based Forensics.} To address the generalization constraints of training from scratch, the field has shifted towards utilizing pre-trained foundation models as sources of universal knowledge. By leveraging higher-ranked semantic priors, these approaches have established highly generalizable baselines across diverse forensic scenarios. Specifically, pre-trained backbones have been successfully adapted for AIGC detection, identifying global synthesis artifacts~\cite{Wen2025fakevlmNIPS, Yan2025aideICLR, Kang2025legionICCV, Yan2025effortICML} and facial manipulations~\cite{Cui2025forensicsAdapterCVPR}. Meanwhile, for Image Forgery Detection and Localization (IFDL), researchers have integrated diverse foundational priors to accurately localize tampering traces in both AIGC inpainting and traditional splicing scenarios~\cite{Huang2025sidaCVPR, Wang2025clue, Peng2025forensicsSAM}. Furthermore, the rich semantic understanding inherent in these models supports interpretable forensics, integrating LLMs to provide detailed textual reasoning for detection verdicts~\cite{huang2024ffaa,Wen2025fakevlmNIPS,Huang2025sidaCVPR}. However, we argue that the widespread dominance of shared upstream encoders (particularly CLIP) establishes a static attack surface, allowing upstream-optimized perturbations to achieve universal anti-forensics attack without accessing specific downstream parameters.

\subsection{Adversarial and Anti-Forensics Attacks} Adversarial strategies in the forensic domain, often termed anti-forensics attack, aim to conceal manipulation traces to evade detection. Since assuming white-box access to model parameters is unrealistic in real-world scenarios, current research focuses on transfer-based black-box strategies. These methods typically leverage model-level surrogates, ranging from standard CNNs~\cite{Hussain2021adversarialWACV} to latent diffusion models~\cite{Zhou2024stealhdiffusionACM}, to optimize transferable perturbations. To further refine image authenticity, recent approaches employ specialized techniques such as frequency analysis~\cite{jia2022frequencyCVPR}, shallow reconstruction~\cite{huang2020fakepolisherACM}, and metric-aware optimization~\cite{Ho2025msgagaarXiv}.

However, these methods face significant limitations in the context of foundation model-based forensics. Primarily, there exists a fundamental surrogate-target mismatch: low-level artifacts captured by conventional surrogates often fail to impact the higher-ranked semantic representations of pre-trained backbones. This limits the transferability of such attacks to advanced detectors. Furthermore, while attacks targeted at VLMs exist~\cite{hu2024prm, fang2024cpgcICCV}, they typically focus on content manipulation (e.g., misclassifying objects) rather than concealing forensic traces. Such semantic distortions often introduce new artifacts that make detection easier. Moreover, these approaches overlook the integrity of interpretable reasoning, failing to address the emerging generation of textual forensic evidence.

\begin{figure*}[t] 
  \centering
  \includegraphics[width=\textwidth]{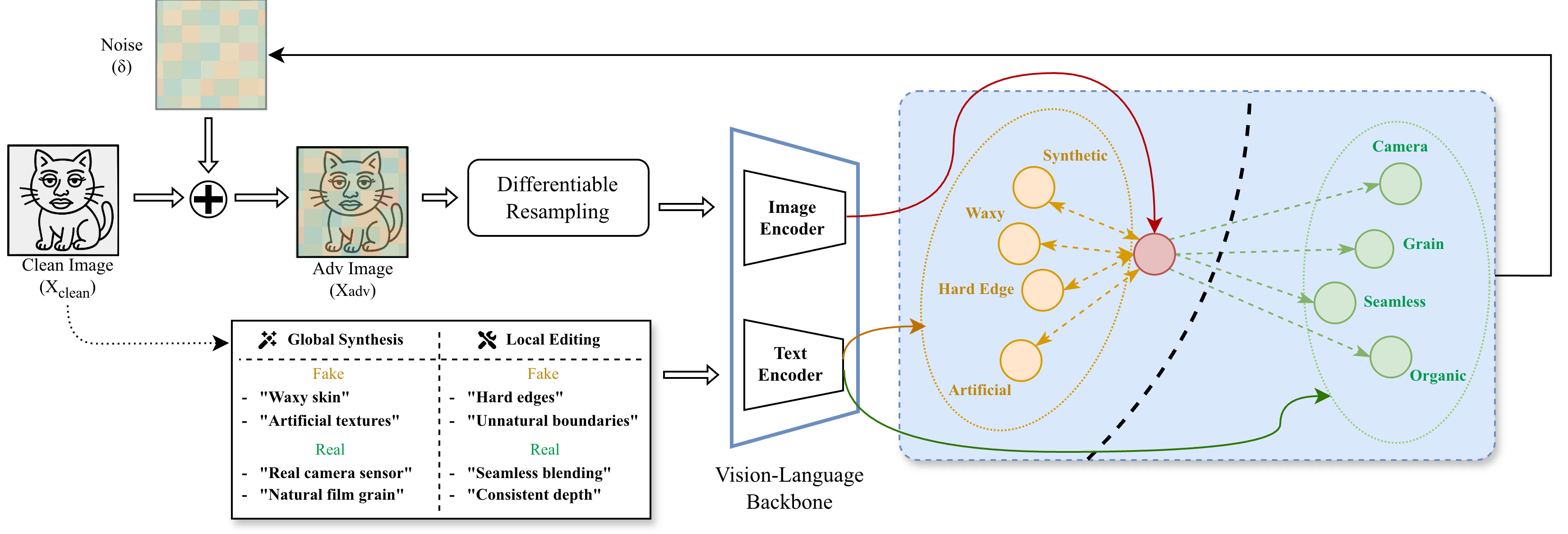} 
\caption{\textbf{Overview of the ForgeryEraser framework.} 
The optimization pipeline incorporates \textbf{Differentiable Resampling} to bridge the resolution gap while suppressing aliasing artifacts. 
Based on a source-aware strategy, the multi-modal guidance loss pulls the image embeddings toward the selected authentic anchors (Green) while pushing them away from forgery anchors (Orange), effectively erasing manipulation traces within the shared feature space.}
\label{fig:framework}
\end{figure*}

\section{Methodology}
\label{sec:method}

\subsection{Problem Formulation and Threat Model}

\textbf{Notation.}
Let $\mathbf{x}$ denote an input image (either real or forged). In this work, we target the CLIP model~\cite{Radford2021clipICML} as our upstream backbone, which comprises an image encoder $E_{img}$ and a text encoder $E_{txt}$. Downstream detectors typically employ $E_{img}$ (often frozen) to extract semantic embeddings $\mathbf{z} = E_{img}(\mathbf{x})$, which are then processed by a learnable head $D_\theta(\cdot)$ to predict the probability of forgery $y = D_\theta(\mathbf{z})$, where $y \to 0$ denotes ``Real'' and $y \to 1$ denotes ``Fake''.

\textbf{Threat Model.}
The attacker has white-box access to the public upstream backbone (including its architecture and gradients), but treats the downstream detector $D_\theta$ as a complete black box. Additionally, reflecting practical scenarios where the forgery creator typically knows how the content was generated, we assume access to a generative prior $s$ indicating whether the target belongs to \textit{global synthesis} or \textit{local editing}. Under this setting, the adversarial objective is to find a perturbation $\boldsymbol{\delta}$ ($||\boldsymbol{\delta}||_\infty \le \epsilon$) such that the detector misclassifies the forged image as authentic: $D_\theta(E_{img}(\mathbf{x} + \boldsymbol{\delta})) \to 0$. Since the gradients of $D_\theta$ are inaccessible, we must exploit the shared upstream backbone to indirectly compromise the unknown downstream detectors.

\subsection{Multi-modal Guidance}

Instead of training surrogate models or maximizing classifier error via Cross-Entropy loss, we directly manipulate the image representation within the semantic feature space of the upstream backbone. Leveraging the aligned text encoder $E_{txt}$, we construct \textit{semantic anchors} that define the direction from forgery to authenticity, and optimize perturbations to guide forged image embeddings toward these authentic anchors.

\textbf{Source-Aware Anchor Construction.} 

We employ a source-aware strategy conditioned on the generative prior $s$ to select domain-specific text prompts. Crucially, we design distinct anchor sets for each scenario, as the semantic cues for authenticity and forgery differ significantly between global synthesis and local editing.
For \textit{Global Synthesis}, we construct \textbf{Global Synthesis Anchors}, comprising authentic prompts $\mathcal{T}_{real}^{global}$ (e.g., ``natural ISO noise'') and forgery prompts $\mathcal{T}_{fake}^{global}$ targeting holistic anomalies (e.g., ``waxy skin'', ``generative artifacts''). For \textit{Local Editing}, we construct \textbf{Local Editing Anchors}, comprising authentic prompts $\mathcal{T}_{real}^{local}$ (e.g., ``seamless blending'') and forgery prompts $\mathcal{T}_{fake}^{local}$ targeting structural discontinuities (e.g., ``unnatural boundaries'', ``hard edges'').
Formally, depending on the generative prior $s \in \{global, local\}$, we select the corresponding prompt sets and encode them into normalized semantic vectors: $\mathcal{A}_{real} = \{ \frac{E_{txt}(t)}{\|E_{txt}(t)\|_2} \mid t \in \mathcal{T}_{real}^{s} \}$ and $\mathcal{A}_{fake} = \{ \frac{E_{txt}(t)}{\|E_{txt}(t)\|_2} \mid t \in \mathcal{T}_{fake}^{s} \}$. This ensures the optimization is strictly guided by domain-specific semantics.

\textbf{Optimization Objective.}
ForgeryEraser optimizes the cosine similarity between image embeddings and the constructed text anchors. To capture both high-level semantics and fine-grained artifacts, we optimize across a set of target layers $\mathcal{S}$. In our default configuration, we set $\mathcal{S}=\{N-1, N\}$ where $N$ denotes the final block. Features from the penultimate layer are projected via CLIP's projection head into the joint text-image space to ensure dimensional alignment.

Let $\mathbf{z}_{adv}^l = E_{img}^l(\mathbf{x} + \boldsymbol{\delta})$ denote the $L_2$-normalized image embedding extracted from layer $l \in \mathcal{S}$ of the adversarial image. We decompose the multi-modal guidance loss into two directional components, where $\cos(\cdot, \cdot)$ denotes the cosine similarity:
\begin{align}
    \mathcal{L}_{pull}^l &= \sum_{\mathbf{a} \in \mathcal{A}_{real}} \left( 1 - \cos(\mathbf{z}_{adv}^l, \mathbf{a}) \right) \\
    \mathcal{L}_{push}^l &= \sum_{\mathbf{a} \in \mathcal{A}_{fake}} \cos(\mathbf{z}_{adv}^l, \mathbf{a})
\end{align}
The total multi-modal guidance loss $\mathcal{L}_{MMG}$ aggregates these components across the selected layers:
\begin{equation}
\label{eq:mmg_loss}
    \mathcal{L}_{MMG} = \sum_{l \in \mathcal{S}} \omega_l \left( \mathcal{L}_{pull}^l + \lambda \cdot \mathcal{L}_{push}^l \right)
\end{equation}
where $\mathcal{L}_{pull}^l$ pulls features towards authentic anchors, while $\mathcal{L}_{push}^l$ repels features from forgery anchors. The hyperparameter $\lambda$ balances these two objectives, and $\omega_l$ assigns weights to different layers (we set $\omega_l = 1/|\mathcal{S}|$ by default).

\subsection{The ForgeryEraser Algorithm}

We integrate the proposed multi-modal guidance loss into a unified optimization framework (Figure~\ref{fig:framework}). To ensure the generated perturbations are robust against input transformations, we incorporate a differentiable preprocessing step.

\textbf{Differentiable Resampling.}
Pre-trained backbones typically require a fixed input resolution (e.g., $224 \times 224$), whereas forensic imagery often retains higher native resolutions. To bridge this gap, we incorporate a differentiable resampling operator $\mathcal{R}(\cdot)$ with anti-aliased interpolation~\cite{zhang2019antialiasICML}, such that the encoder input becomes $\mathcal{R}(\mathbf{x} + \boldsymbol{\delta})$. Backpropagating through $\mathcal{R}$ allows optimizing the high-resolution perturbation $\boldsymbol{\delta}$ while suppressing aliasing artifacts, enhancing robustness to preprocessing.

\textbf{Optimization Loop.}
We employ the Momentum Iterative Fast Gradient Sign Method (MI-FGSM)~\cite{dong2018mifgsmCVPR} to stabilize the update trajectory. Let $\boldsymbol{\delta}_t$ denote the perturbation at iteration $t \in \{1, \ldots, T\}$ and $\mathbf{g}_t$ the accumulated momentum:
\begin{align}
\mathbf{g}_{t+1} &= \mu \cdot \mathbf{g}_t + \frac{\nabla_{\boldsymbol{\delta}} \mathcal{L}_{MMG}(\mathcal{R}(\mathbf{x} + \boldsymbol{\delta}_t))}{||\nabla_{\boldsymbol{\delta}} \mathcal{L}_{MMG}(\mathcal{R}(\mathbf{x} + \boldsymbol{\delta}_t))||_1} \\
\boldsymbol{\delta}_{t+1} &= \text{Clip}_{\epsilon} \left( \boldsymbol{\delta}_t - \alpha \cdot \text{sign}(\mathbf{g}_{t+1}) \right)
\end{align}
where $\mu$ is the decay factor and $\alpha$ is the step size. The operation $\text{Clip}_{\epsilon}$ constrains the perturbation within the $L_\infty$-ball $[-\epsilon, \epsilon]$. We apply gradient subtraction to minimize $\mathcal{L}_{MMG}$, guiding the representation towards the authentic anchors. After $T$ iterations, the final adversarial example is clamped to the valid pixel range: $\mathbf{x}_{adv} = \text{Clip}_{[0,1]}(\mathbf{x} + \boldsymbol{\delta}_{T})$.

\begin{table*}[t]
\centering
\small 
\setlength{\tabcolsep}{3pt} 

\caption{\textbf{Universal Anti-Forensics Attack Performance.} 
Detection accuracy (\%) and Relative Accuracy Change ($\mathcal{R}_{\Delta}$) on six AIGC detectors under perturbation budgets $\epsilon \in \{4/255, 8/255\}$. 
Negative $\mathcal{R}_{\Delta}$ values indicate a reduction in detection accuracy (attack success). 
\textit{Note:} For SIDA, Real accuracy is shared across subsets.}

\label{tab:main_results}

\begin{tabular}{ll c rr c rr c c rr c rr}
\toprule

\multirow{3.5}{*}{\textbf{Target Model}} & \multirow{3.5}{*}{\textbf{Dataset}} & 
\multicolumn{6}{c}{\textbf{Real Images}} & \multicolumn{1}{c}{} & 
\multicolumn{6}{c}{\textbf{Fake Images}} \\
\cmidrule(lr){3-8} \cmidrule(lr){10-15}

& & \multirow{2.5}{*}{\textbf{Clean}} & \multicolumn{2}{c}{\textbf{$\epsilon=4$}} & \multicolumn{1}{c}{} & \multicolumn{2}{c}{\textbf{$\epsilon=8$}} & & 
\multirow{2.5}{*}{\textbf{Clean}} & \multicolumn{2}{c}{\textbf{$\epsilon=4$}} & \multicolumn{1}{c}{} & \multicolumn{2}{c}{\textbf{$\epsilon=8$}} \\
\cmidrule(lr){4-5} \cmidrule(lr){7-8} \cmidrule(lr){11-12} \cmidrule(lr){14-15}

& & & Acc & \multicolumn{1}{c}{$\mathcal{R}_{\Delta}(\%)$} & & Acc & \multicolumn{1}{c}{$\mathcal{R}_{\Delta}(\%)$} & & 
& Acc & \multicolumn{1}{c}{$\mathcal{R}_{\Delta}$(\%)} & & Acc & \multicolumn{1}{c}{$\mathcal{R}_{\Delta}(\%)$} \\
\midrule


\multirow{2}{*}{SIDA \scriptsize{(CVPR'25)}} & \scriptsize{SID-Set(FullSync)} & 
\multirow{2}{*}{95.3} & \multirow{2}{*}{\textbf{99.4}} & \multirow{2}{*}{+4.3} & & \multirow{2}{*}{\textbf{99.1}} & \multirow{2}{*}{+3.9} & & 
99.5 & \textbf{47.0} & -52.8 & & \textbf{26.5} & -73.0 \\

& \scriptsize{SID-Set(Tampered)} & & & & & & & & 
93.7 & \textbf{12.0} & -87.1 & & \textbf{10.0} & -89.3 \\

\midrule

AIDE \scriptsize{(ICLR'25)} & \scriptsize{AIGCDetectBenchmark} & 
95.1 & \textbf{98.4} & +3.5 & & \textbf{99.4} & +4.5 & & 
96.5 & \textbf{31.6} & -67.3 & & \textbf{14.2} & -85.2 \\

\midrule

FakeVLM \scriptsize{(NIPS'25)} & \scriptsize{FakeClue} & 
97.3 & 97.0 & -0.3 & & \textbf{97.5} & +0.2 & & 
99.3 & \textbf{55.6} & -44.0 & & \textbf{37.6} & -62.2 \\

\midrule

LEGION \scriptsize{(ICCV'25)} & \scriptsize{UniversalFakeDetect} & 
98.8 & \textbf{99.5} & +0.7 & & \textbf{99.9} & +1.1 & & 
74.7 & \textbf{4.8} & -93.6 & & \textbf{0.5} & -99.4 \\

\midrule

Effort \scriptsize{(ICML'25)} & \scriptsize{Protocol-1} & 
67.2 & \textbf{97.9} & +45.6 & & \textbf{95.5} & +42.1 & & 
90.8 & \textbf{9.1} & -89.9 & & \textbf{9.4} & -89.7 \\

\midrule

Forensics Adapter \scriptsize{(CVPR'25)} & \scriptsize{Protocol-1} & 
93.1 & \textbf{97.8} & +5.1 & & \textbf{97.7} & +5.0 & & 
62.5 & \textbf{5.5} & -91.1 & & \textbf{5.6} & -91.0 \\

\bottomrule

\end{tabular}
\end{table*}

\section{Experiments}
\label{sec:exp}

\subsection{Experimental Setup}

\textbf{Victim Models.} To validate the systemic vulnerability of foundation model-based forensics, we select six representative state-of-the-art AIGC detectors that incorporate CLIP or its variants (e.g., OpenCLIP) as their upstream visual encoder. Our selection encompasses SIDA~\cite{Huang2025sidaCVPR}, AIDE~\cite{Yan2025aideICLR}, FakeVLM~\cite{Wen2025fakevlmNIPS}, LEGION~\cite{Kang2025legionICCV}, Effort~\cite{Yan2025effortICML}, and Forensics Adapter~\cite{Cui2025forensicsAdapterCVPR}. We utilize official pre-trained weights for all models. For LEGION, we followed the official protocol to fine-tune the classification head on the ProGAN~\cite{gao2019proganKDD} dataset.

\textbf{Datasets and Evaluation Tasks.} To demonstrate the unified efficacy of our attack, we conduct evaluations on official benchmarks categorized into two primary regimes. 
\textbf{1) Global Synthesis Detection:} Targeting fully synthetic images, we evaluate AIDE on AIGCDetectBenchmark~\cite{zhong2024aigcbench}, FakeVLM on FakeClue~\cite{Wen2025fakevlmNIPS}, and LEGION on UniversalFakeDetect (UFD)~\cite{Ojha2023universalCVPR}. Additionally, we evaluate SIDA on the \textit{Full-Synthesis} subset of SID-Set~\cite{Huang2025sidaCVPR}.
\textbf{2) Local Editing Detection:} Targeting locally edited content, we adopt the rigorous Protocol-1 (Cross-Dataset Evaluation)~\cite{Yan2025effortICML} to evaluate Effort and Forensics Adapter on Deepfake benchmarks. Furthermore, we evaluate SIDA on the \textit{Tampered} subset of SID-Set, representing AIGC inpainting scenarios. This ensures that each model is attacked within the specific domain and dataset environment of its original benchmark.

\textbf{Evaluation Metrics.} We report the Classification Accuracy on Clean images ($\text{Acc}_{clean}$) to establish baseline performance. To quantify the attack impact, we calculate the Relative Accuracy Change ($\mathcal{R}_{\Delta}$), defined as $(\text{Acc}_{adv} - \text{Acc}_{clean}) / \text{Acc}_{clean} \times 100\%$. This metric represents the percentage change relative to the baseline. A negative $\mathcal{R}_{\Delta}$ indicates a reduction in detection accuracy (successful anti-forensics attack), whereas a positive value indicates improved detection.

\textbf{Implementation Details.} 
We integrate ForgeryEraser into the MI-FGSM~\cite{dong2018mifgsmCVPR} optimizer with a momentum decay of $\mu=1.0$ and set the loss balancing hyperparameter $\lambda=1.0$. All experiments are conducted on NVIDIA A100 GPUs. We evaluate effectiveness under two perturbation budgets: a low-budget setting ($\epsilon=4/255, T=4, \alpha=1/255$) and a standard setting ($\epsilon=8/255, T=8, \alpha=2/255$).
Adhering to our source-aware strategy, we dynamically assign domain-specific anchors based on the generative source of the target dataset. For benchmarks involving \textit{Global Synthesis}, we optimize using the Global Synthesis Anchors ($\mathcal{T}^{global}$). Conversely, for benchmarks involving \textit{Local Editing}, we employ the Local Editing Anchors ($\mathcal{T}^{local}$). The complete list of text prompts is provided in the Appendix.

\subsection{Main Results}

\textbf{Quantitative Analysis.}
Table~\ref{tab:main_results} reports the performance of ForgeryEraser across all evaluated detectors. Under the standard perturbation budget ($\epsilon=8/255$), the attack achieves substantial success rates, driving detection accuracy down to single digits for multiple architectures—most notably reducing LEGION to 0.5\% and Forensics Adapter to 5.6\%. Even models designed for high generalization, such as Effort and AIDE, suffer relative accuracy drops ($\mathcal{R}_{\Delta}$) exceeding 85\%. Crucially, the attack maintains effectiveness under the low-budget setting ($\epsilon=4/255$); for instance, accuracy on SIDA (targeting Local Editing) drops by over 87\%. These results demonstrate that exploiting the shared upstream backbone enables universal anti-forensics attack against diverse downstream AIGC detectors without accessing their specific parameters.

\textbf{Semantic Refinement on Real Images.}
Beyond anti-forensics attack, Table~\ref{tab:main_results} highlights a notable phenomenon we term \textit{Semantic Refinement}—where perturbations enhance authentic traits in Real images. We observe a consistent performance boost on Real images across multiple evaluators. This effect is most pronounced in Effort, where detection accuracy improves from a baseline of 67.2\% to 95.5\% under attack.
This empirical evidence suggests that ForgeryEraser induces a directed semantic migration. Rather than injecting random noise to merely cross a decision boundary, ForgeryEraser actively guides features to encode authentic attributes defined by the backbone (e.g., natural textures). This alignment effectively brings the representation closer to the backbone's definition of ``Real'' than the original image itself.

\begin{table}[t]
\centering
\caption{\textbf{Cross-Generator Generalization.} 
Detection accuracy (\%) of LEGION on specific subsets of the UFD dataset, covering both Diffusion and GAN architectures. }
\label{tab:gen_breakdown}
\resizebox{\linewidth}{!}{
\begin{tabular}{l cc | l cc} 
\toprule
\multicolumn{3}{c|}{\textbf{Diffusion Models}} & \multicolumn{3}{c}{\textbf{GAN Models}} \\ 
\textbf{Generator} & \textbf{Clean (\%)} & \textbf{Adv. (\%)} & \textbf{Generator} & \textbf{Clean (\%)} & \textbf{Adv. (\%)} \\
\midrule
LDM-100      & 94.8 & \textbf{0.3} & ProGAN    & 100.0 & \textbf{1.1} \\
LDM-200      & 92.5 & \textbf{0.6} & GauGAN    & 100.0 & \textbf{0.9} \\
LDM-200-CFG  & 70.6 & \textbf{0.3} & CycleGAN  & 99.9  & \textbf{0.6} \\
GLIDE-100-10 & 70.2 & \textbf{0.1} & StarGAN   & 99.9  & \textbf{1.2} \\
GLIDE-100-27 & 71.3 & \textbf{0.0} & BigGAN    & 76.8  & \textbf{0.4} \\
GLIDE-50-27  & 72.2 & \textbf{0.1} & StyleGAN  & 99.9  & \textbf{1.2} \\
DALL-E       & 90.4 & \textbf{0.4} & StyleGAN2 & 76.8  & \textbf{0.4} \\
\midrule
\rowcolor{GrayLine} \textbf{Avg.} & 81.9 & \textbf{0.25} & \textbf{Avg.} & 91.6 & \textbf{0.76} \\
\bottomrule
\end{tabular}
}
\end{table}

\textbf{Cross-Generator Generalization.} 
As shown in Table~\ref{tab:gen_breakdown}, ForgeryEraser demonstrates consistent effectiveness across diverse generative architectures within the Global Synthesis domain. When evaluating LEGION on expanded subsets of the UFD dataset, accuracy drops to near-zero levels for both Diffusion-based (0.25\%) and GAN-based (0.76\%) images. 
This consistency highlights a key advantage of our approach. Unlike traditional anti-forensic methods that often overfit to model-specific pixel artifacts (e.g., GAN fingerprints), ForgeryEraser targets the shared high-level inconsistencies inherently captured by the CLIP backbone. By guiding forged embeddings towards authentic anchors, ForgeryEraser achieves universal anti-forensics attack, effectively bridging the gap between adversarial training (GANs) and iterative denoising (Diffusion).

\subsection{Analysis \& Visualization}

\begin{figure}[t]
  \centering
  \includegraphics[width=0.95\linewidth]{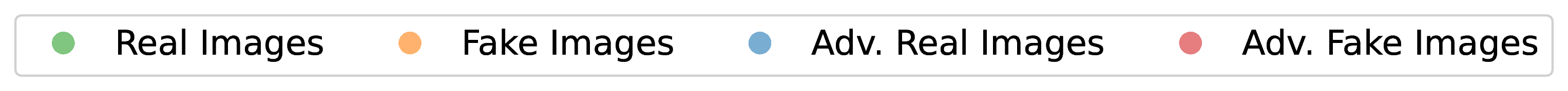} 
  
  \par 
  
\begin{minipage}[t]{0.48\linewidth} 
    \centering
    \includegraphics[width=\linewidth]{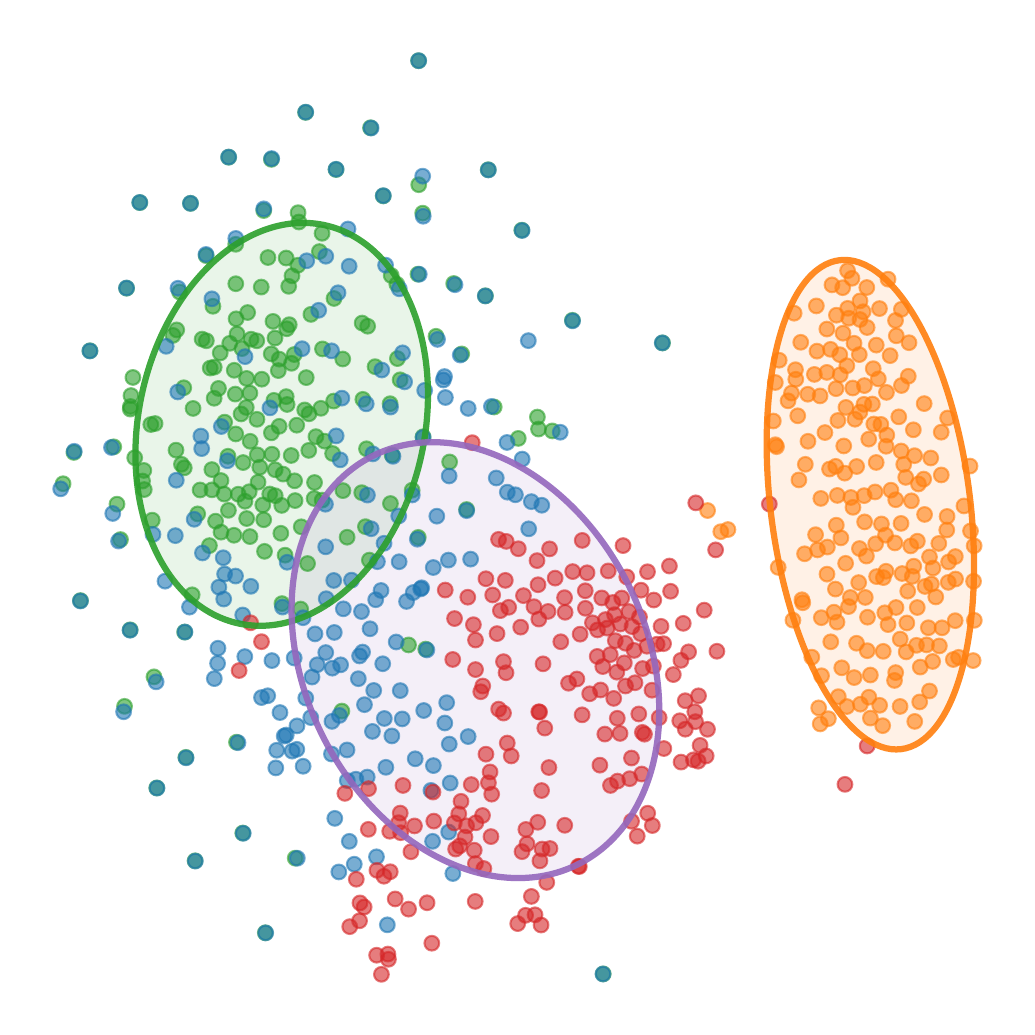}
    \centerline{\small (a) Dog}
  \end{minipage}
  \hfill
  {\color{gray}\vrule width 0.5pt}%
  \hfill
  \begin{minipage}[t]{0.48\linewidth}
    \centering
    \includegraphics[width=\linewidth]{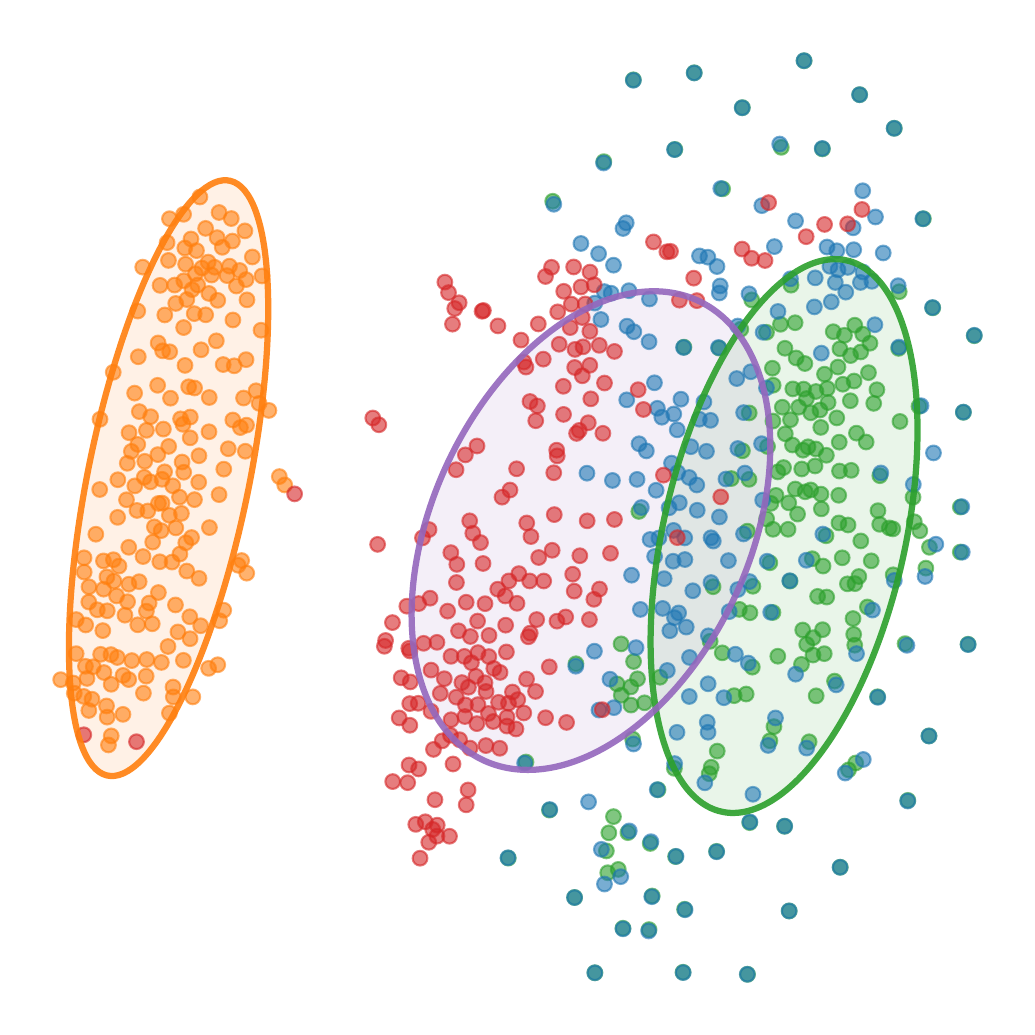}
    \centerline{\small (b) Cat}
  \end{minipage}
  
\caption{\textbf{Feature Space Visualization (t-SNE).} 
Projections of CLIP embeddings for \textbf{(a)} Dog and \textbf{(b)} Cat samples from the ProGAN subset, visualizing Real and Fake images before and after the attack.}
  \label{fig:tsne_vis}
\end{figure}

\textbf{Feature Space Shift.}
To visualize the mechanism of ForgeryEraser, we project the CLIP image embeddings into a 2D plane using t-SNE. We select two distinct semantic categories, Dog and Cat, from the ProGAN subset of UFD~\cite{gao2019proganKDD} to demonstrate that our attack targets authenticity independent of object content. Figure~\ref{fig:tsne_vis} visualizes the distributions of Real, Fake, and attacked samples.

As observed in Figure~\ref{fig:tsne_vis}, the pre-trained CLIP backbone inherently separates Real and Fake images into distinct clusters. This separation—persisting across different objects—confirms that the backbone captures a latent authentic semantic direction.
Upon applying ForgeryEraser, two critical phenomena emerge:
\textit{1) Decoupling and Migration:} The adversarial Fake embeddings (Red) structurally decouple from the original Fake distribution. Instead of scattering randomly, they migrate directionally towards the Real cluster. This qualitative shift is confirmed by quantitative evidence: for both Dog and Cat subsets, the detection accuracy for Fake images collapses from 100.0\% to 1.5\% and 0.0\% respectively. This confirms that the visual migration corresponds to a successful anti-forensics attack.
\textit{2) Synchronized Alignment:} Crucially, the adversarial Real images do not remain static but instead exhibit a synchronized migration with the adversarial Fake images toward a shared target distribution. This convergence indicates that our attack guides both sets of embeddings into a specific semantic region that downstream detectors universally recognize as ``Real''. Quantitatively, the detection accuracy for Real images remains robust at 100.0\%, confirming that ForgeryEraser preserves the semantic integrity of authentic content while effectively erasing forgery traces.

\begin{figure*}[t] 
  \centering
  \includegraphics[width=\linewidth]{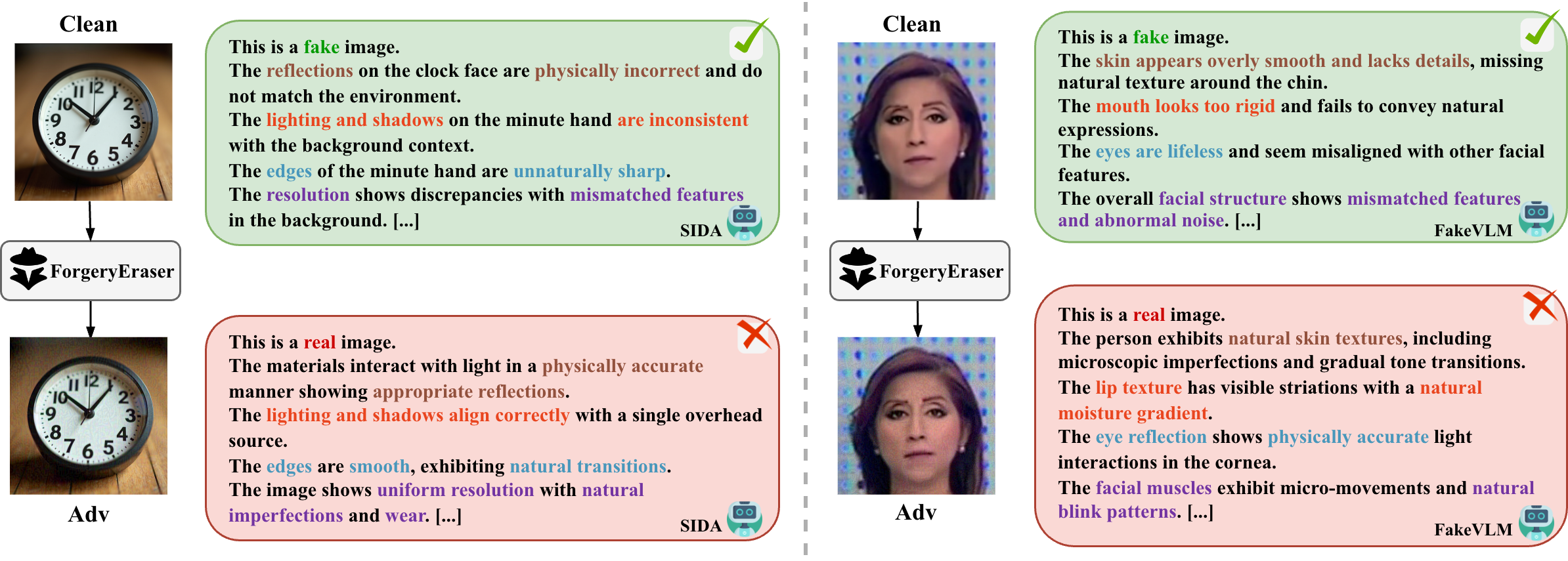}
  
\caption{\textbf{Manipulating Interpretability on SIDA (Left) and FakeVLM (Right).} 
\textbf{Top Row:} Detectors correctly localize and describe visual artifacts on clean images. 
\textbf{Bottom Row:} Under the ForgeryEraser attack, models are induced to fabricate justifications for authenticity. Note that matching text colors across rows highlight opposing descriptions generated for identical visual features before and after the attack.}
  \label{fig:interpretability}
\end{figure*}

\textbf{Manipulating Interpretability.}
Beyond detection accuracy, ForgeryEraser induces explainable forensic models to fabricate authentic justifications for forged images. We conduct a qualitative study on SIDA~\cite{Huang2025sidaCVPR} and FakeVLM~\cite{Wen2025fakevlmNIPS} to analyze how the attack influences generated explanations. Figure~\ref{fig:interpretability} presents the results on a generated clock image (SIDA) and a Deepfake face (FakeVLM).

The results reveal that ForgeryEraser causes detectors to invert their verdicts and fabricate plausible justifications. In the SIDA case (Left), the model initially correctly identifies artifacts such as ``physically incorrect reflections.'' However, under ForgeryEraser, these descriptions are effectively overwritten. The system generates fabricated justifications for authenticity, citing ``physically accurate light interactions'' which directly contradict the visual artifacts. 
Similarly, for the Deepfake face evaluated on FakeVLM (Right), the interpretation shifts from flagging ``lifeless eyes'' to describing ``natural moisture gradients.'' 
This indicates that ForgeryEraser is not limited to confusing the decision boundary; by guiding forged embeddings toward authentic anchors in the shared backbone, it effectively induces downstream models to fabricate plausible justifications for forged content, regardless of the specific attributes they are designed to detect.

\begin{figure}[t] 
  \centering
  \includegraphics[width=\linewidth]{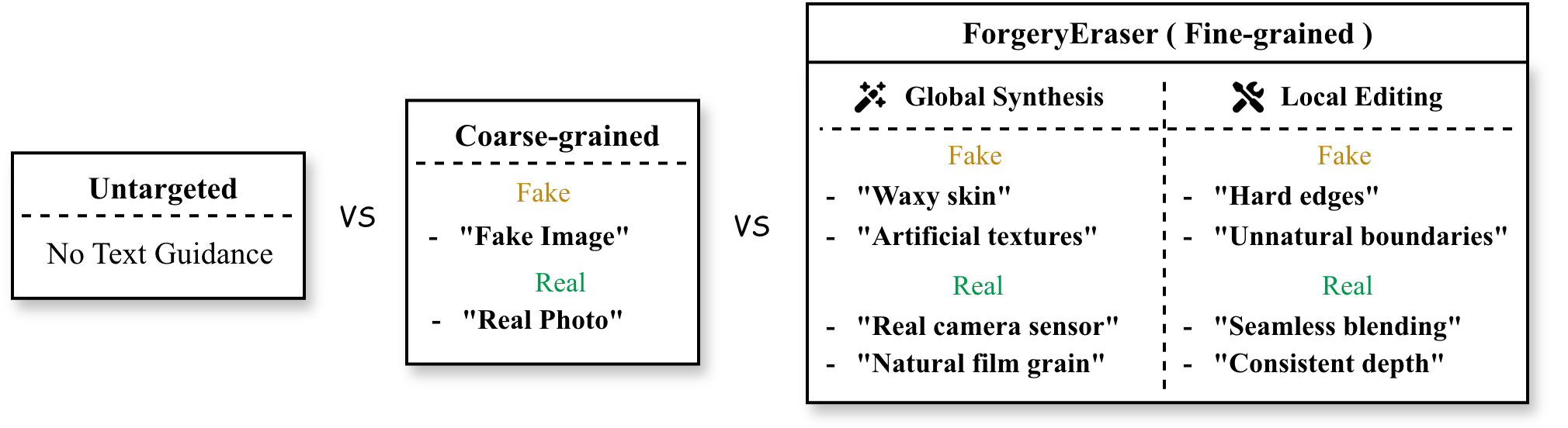} 
  
  
  \caption{\textbf{Visualization of Semantic Anchors.} 
  Comparison of text guidance strategies with varying granularities: 
  \textbf{Untargeted} (no text guidance), 
  \textbf{Coarse-grained} (generic class labels), and 
  \textbf{ForgeryEraser} (fine-grained attribute descriptions defined by the source-aware strategy).}
  \label{fig:ablation_anchors}
  \label{fig:prompt_comparison}
\end{figure}

\subsection{Ablation Studies}

We conduct ablation studies to validate the necessity of each component within ForgeryEraser, specifically examining the impact of text guidance granularity and the effectiveness of source-aware anchors.

\textbf{Impact of Semantic Granularity.}
To isolate the contribution of the semantic text encoder, we compare ForgeryEraser against two baselines (visualized in Figure~\ref{fig:prompt_comparison}). 
First, Untargeted Disruption, inspired by adversarial noise strategies, aims to disrupt the image representation by maximizing feature divergence from the input without a specific directional target. This undirected disruption risks inducing random semantic distortions that can produce the opposite effect—making forged images even easier to detect (as evidenced by Effort in Table~\ref{tab:ablation_component}). 
Second, Coarse-grained Guidance utilizes the multi-modal guidance loss but relies on generic class labels (e.g., ``a real photo'') rather than fine-grained attribute descriptions.
As shown in Table~\ref{tab:ablation_component}, the results reveal a clear performance ranking. Untargeted disruption yields the lowest success rates. While Coarse-grained Guidance improves performance by introducing basic directionality, it lacks optimization precision. In contrast, ForgeryEraser achieves superior performance. By explicitly modeling specific artifacts (e.g., ``waxy skin'') versus authentic traits, ForgeryEraser precisely guides features into the ``Real'' distribution. This confirms that attribute-rich guidance is essential for the feature-level surrogate, providing the directional precision required to navigate the shared feature space without parametric training.

\begin{table}[t]
\centering
\setlength{\tabcolsep}{4pt} 
\caption{\textbf{Component Ablation Study.} 
Comparison of anti-forensics attack performance across different semantic guidance granularities. 
The table reports detection accuracy (\%) on Real and Fake images for four victim models.}
\label{tab:ablation_component}
\resizebox{\linewidth}{!}{
\begin{tabular}{l|cc|cc|cc|cc}
\toprule
\multirow{2}{*}{\textbf{Method}} & \multicolumn{2}{c|}{\textbf{SIDA}} & \multicolumn{2}{c|}{\textbf{LEGION}} & \multicolumn{2}{c|}{\textbf{FakeVLM}} & \multicolumn{2}{c}{\textbf{Effort}} \\
& Real & Fake & Real & Fake & Real & Fake & Real & Fake \\
\midrule
Baseline & 95.3 & 93.7 & 98.8 & 74.7 & 97.3 & 99.3 & 67.2 & 90.8 \\
\midrule
Untargeted & 78.4 & 62.6 & 88.2 & 34.5 & 77.3 & 80.7 & 41.1 & 91.4 \\
Coarse-grained & \underline{97.4} & \underline{25.0} & \underline{99.6} & \underline{3.5} & \underline{91.0} & \underline{54.9} & \underline{94.1} & \underline{12.9} \\
 \textbf{ForgeryEraser} & \textbf{99.1} & \textbf{10.0} & \textbf{99.9} & \textbf{0.5} & \textbf{97.5} & \textbf{37.6} & \textbf{95.5} & \textbf{9.4} \\
\bottomrule
\end{tabular}
}
\end{table}

\begin{figure}[h]
  \centering
 
  \includegraphics[width=\linewidth]{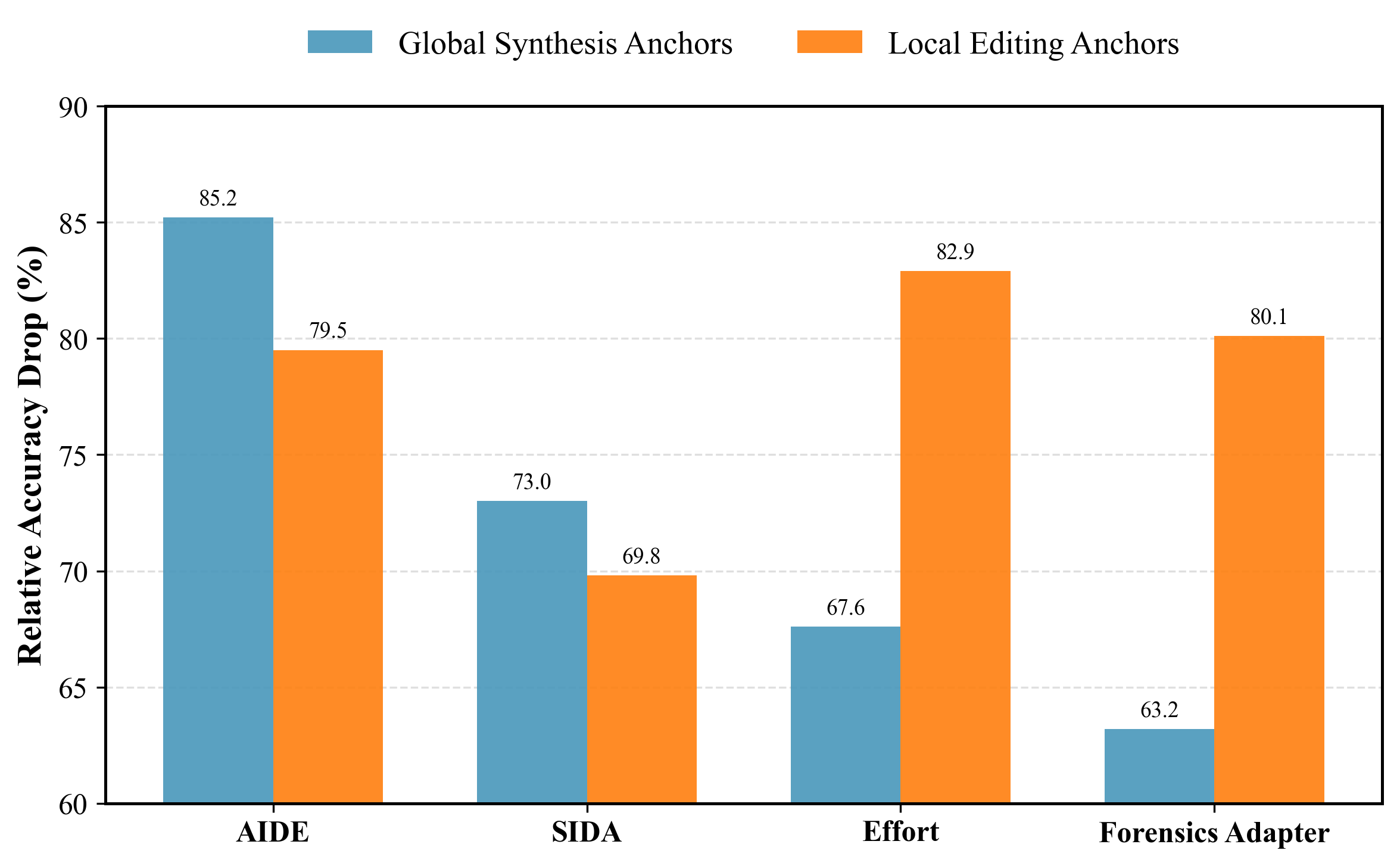}
  \caption{\textbf{Validation of source-awareness strategy.} Comparison of attack performance (Relative Accuracy Drop) when using \textbf{Matched} anchors (e.g., Global Synthesis Anchors applied to Global Synthesis images) versus \textbf{Mismatched} anchors.}
  \label{fig:cross_anchor}
\end{figure}

\textbf{Validation of Source-Awareness.}
To verify the necessity of our source-aware strategy, we cross-evaluate victim models using matched versus mismatched anchors (Figure~\ref{fig:cross_anchor}). We categorize models into Global Synthesis targets (AIDE, SIDA) and Local Editing targets (Effort, Forensics Adapter).
We observe a distinct domain preference. While mismatched anchors achieve moderate anti-forensics attack (indicating some shared vulnerability in the backbone), the performance gap is significant. Specifically, Local Editing detectors are far more susceptible to Local Editing Anchors, whereas Global Synthesis detectors achieve optimal results only under their matched Global Synthesis Anchors. This confirms that while the shared backbone is universal, downstream detectors specialize in distinct artifact domains. Thus, maximizing attack efficacy requires aligning the adversarial anchors with the generative source governing the target detector.

\subsection{Robustness Evaluation}

\begin{figure}[t]
  \centering
  \includegraphics[width=\linewidth]{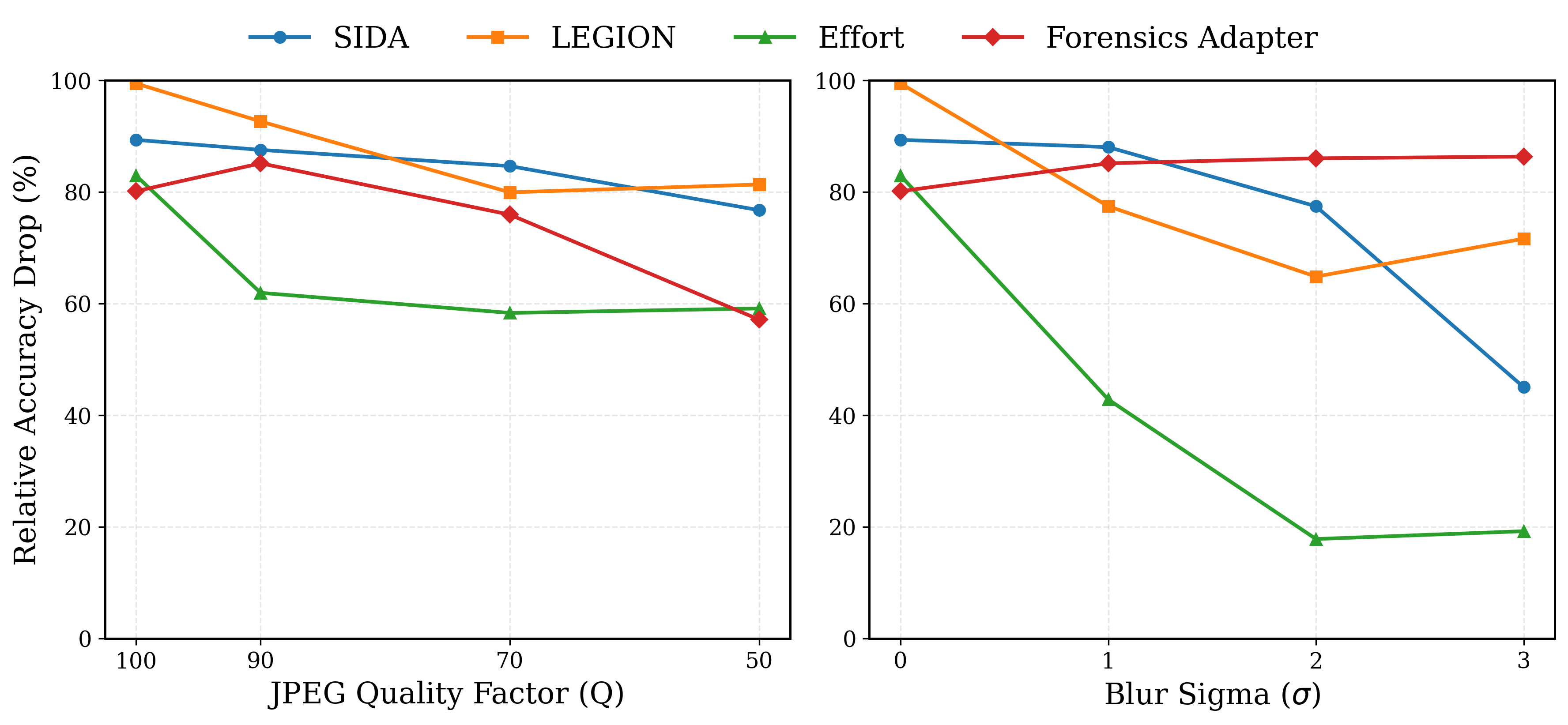}
  \caption{\textbf{Robustness against Image Distortions.} 
  Attack performance measured by Relative Accuracy Change ($\mathcal{R}_{\Delta}$) under varying levels of \textbf{JPEG Compression} (Left) and \textbf{Gaussian Blur} (Right).}
  \label{fig:robustness}
\end{figure}

To evaluate the robustness of our anti-forensics attack against common image distortions, we subject adversarial examples to JPEG compression ($Q \in \{100,90,70,50\}$) and Gaussian Blur ($\sigma \in \{0,1,2,3\}$). We exclude AIDE and FakeVLM from this analysis due to their baseline instability on distorted data. Preliminary experiments reveal that these two detectors exhibit severe prediction biases on low-quality inputs, often defaulting to a single class regardless of actual authenticity, which renders adversarial evaluation invalid. This limits our ability to draw robust conclusions about the attack's performance under these conditions.

As shown in Figure~\ref{fig:robustness}, ForgeryEraser demonstrates strong structural robustness. Under JPEG compression, the anti-forensics attack efficacy (measured by $\mathcal{R}_{\Delta}$) remains stable across all models, maintaining substantial performance even at the aggressive compression rate of $Q=50$. Similarly, under Gaussian Blur, the attack maintains stable effectiveness across most detectors.

\textbf{Mechanism of Robustness.} We attribute this resilience to two key factors:
1) Frequency-Aware Optimization: The Differentiable Resampling strategy explicitly suppresses fragile high-frequency noise during optimization. This forces the perturbation to encode information into robust, low-frequency structural bands that survive compression.
2) Semantic Stability: Unlike traditional attacks that manipulate fragile pixel-level statistical residuals, ForgeryEraser operates within the semantic feature space inherited from the upstream backbone. By embedding ``authentic'' concepts (e.g., natural textures) into the image content itself, the adversarial features become intrinsic to the visual representation, making them significantly harder to remove via standard filtering than superficial noise.

\textit{Note on Effort:} We observe a decline for Effort under severe blur ($\sigma \ge 2.0$). However, tests on clean data reveal that this stems from the model's baseline sensitivity: its accuracy on clean Real images drops significantly under such blur. This severe baseline degradation limits the measurable margin for adversarial impact, rather than indicating a failure of the attack mechanism itself.

\section{Conclusion}
\label{sec:conclusion}

In this paper, we reveal a systemic vulnerability in modern AIGC forensics arising from the widespread reliance on shared upstream backbones (e.g., CLIP). We propose \textbf{ForgeryEraser}, a universal anti-forensics attack framework against AIGC detectors that exploits this vulnerability by directly manipulating the inherited representations within the accessible upstream backbone. Leveraging the multi-modal guidance loss, ForgeryEraser drives forged embeddings toward text-derived authentic anchors while repelling them from forgery anchors, effectively erasing manipulation traces within the shared feature space.
Extensive experiments demonstrate that ForgeryEraser causes substantial performance degradation to advanced AIGC detectors on both Global Synthesis and Local Editing benchmarks across diverse generative architectures, without accessing their specific parameters. Notably, ForgeryEraser induces explainable forensic models to generate explanations consistent with authentic images for forged images. Furthermore, robustness experiments confirm that these semantic perturbations survive common image distortions, validating their practical applicability.
This work highlights the need for the forensic community to reconsider the widespread reliance on shared upstream backbones and to develop next-generation systems resilient to semantic-level manipulation.

\section*{Impact Statement}

This paper aims to advance the security of Machine Learning in the domain of digital media forensics. We reveal a critical systemic risk where the widespread reliance on shared upstream backbones (e.g., CLIP) creates a vulnerability prone to universal anti-forensics attack. While ForgeryEraser demonstrates the capability to cause substantial performance degradation to state-of-the-art Global Synthesis and Local Editing detectors, identifying this systemic vulnerability is a necessary prerequisite for developing resilient defenses. By demonstrating that detectors can be deceived not just in their verdicts but also in their interpretable reasoning, our work provides the theoretical and empirical basis required to design robust forensic systems. We emphasize that responsible disclosure and awareness of these vulnerabilities are essential for maintaining trust in an era increasingly dominated by AI-generated content.


\bibliographystyle{icml2026}

\newpage
\appendix
\onecolumn

\renewcommand{\arraystretch}{1}

\section{Target Model Specifications}
\label{app:backbones}
Table~\ref{tab:backbone_transpose} details the architectural specifications of the target detectors. The data confirms that 5 out of 6 state-of-the-art models rely on variants of the same CLIP Vision Transformer (ViT)-L/14 architecture. This widespread reliance on shared upstream backbones enables the universal transferability of ForgeryEraser.

\begin{table}[!ht]
\centering
\caption{\textbf{Target Model Specifications.} 
Architectural details of the six evaluated AIGC detectors. 
The table lists the upstream visual backbone, pre-training source, and input resolution required by each model.}
\label{tab:backbone_transpose}
\small

\begin{tabular*}{0.8\textwidth}{@{\extracolsep{\fill}}l | cccccc}
\toprule
\textbf{Attribute} & \textbf{SIDA} & \textbf{Effort} & \textbf{Forensics Adapter} & \textbf{FakeVLM} & \textbf{LEGION} & \textbf{AIDE} \\
\midrule
\textbf{Backbone} & ViT-L/14 & ViT-L/14 & ViT-L/14 & ViT-L/14 & ViT-L/14 & ConvNeXt \\
\textbf{Source} & OpenAI & OpenAI & OpenAI & OpenAI & OpenAI & OpenCLIP \\
\textbf{Resolution} & $224 \times 224$ & $224 \times 224$ & $224 \times 224$ & $336 \times 336$ & $336 \times 336$ & $256 \times 256$ \\
\bottomrule
\end{tabular*}
\end{table}

\section{Detailed Dataset Composition}
\label{app:dataset_details}

Tables~\ref{tab:sida_details} through \ref{tab:legion_details} detail the composition of our evaluation benchmarks. The test samples span diverse generative families including GANs (e.g., ProGAN, StyleGAN) and Diffusion Models (e.g., Stable Diffusion, Midjourney). This broad coverage validates that our source-aware anchors are effective across diverse synthesis paradigms.

\begin{table}[!ht]
\centering
\caption{\textbf{Statistics of SID-Set.} The dataset contains authentic samples and two forgery types: Local Editing (AIGC inpainting) and Global Synthesis (Full Synthesis).}

\label{tab:sida_details}
\small

\begin{tabular*}{0.6\textwidth}{c|c|l@{\extracolsep{\fill}}c}
\toprule
\textbf{Model} & \textbf{Dataset} & \textbf{Category (Type)} & \textbf{Count} \\
\midrule
\multirow{5}{*}{\textbf{SIDA}} & \multirow{5}{*}{\textbf{SID-Set}}
& Authentic (Real) & 20,000 \\
& & Tampered (Fake) & 20,000 \\
& & Fully Synthetic (Fake) & 20,000 \\
\cmidrule{3-4}
& & \textbf{Total} & \textbf{60,000} \\
\bottomrule
\end{tabular*}
\end{table}

\begin{table}[!ht]
\centering
\caption{\textbf{Statistics of FakeClue Benchmark.} The benchmark includes diverse domains (e.g., Documents, Satellite) in addition to standard face and object synthesis subsets.}

\label{tab:fakevlm_details}
\small
\begin{tabular*}{0.8\textwidth}{c|c|l@{\extracolsep{\fill}}ccc}
\toprule
\textbf{Model} & \textbf{Dataset} & \textbf{Subset} & \textbf{Real} & \textbf{Fake} & \textbf{Total} \\
\midrule
\multirow{6}{*}{\textbf{FakeVLM}} & \multirow{6}{*}{\textbf{FakeClue}}
& Chameleon & - & 441 & 441 \\
& & Doc & 116 & 460 & 696 \\
& & FaceForensics++ & 236 & 932 & 1168 \\
& & GenImage (Agg.) & 1024 & 916 & 1940 \\
& & Satellite & 432 & 443 & 875 \\
\cmidrule{3-6}
& & \textbf{Total} & \textbf{1,808} & \textbf{3,192} & \textbf{5,000} \\
\bottomrule
\end{tabular*}
\end{table}

\begin{table}[!ht]
\centering
\caption{\textbf{Statistics of Protocol-1 (Cross-Dataset).} An aggregation of seven Deepfake face datasets (totaling $>125k$ images) for cross-dataset evaluation.}

\label{tab:protocol1_details}
\small
\begin{tabular*}{0.8\textwidth}{c|c|l@{\extracolsep{\fill}}ccc}
\toprule
\textbf{Model} & \textbf{Dataset} & \textbf{Dataset Subset} & \textbf{Real} & \textbf{Fake} & \textbf{Total} \\
\midrule
\multirow{8}{*}{\shortstack{\textbf{Effort} \\\&\\ \textbf{Forensic Adapter}}} & \multirow{8}{*}{\textbf{Protocol-1}}
& CDF & 3,560 & 6,800 & 10,360 \\
& & DFDCP & 5,280 & 9,680 & 14,960 \\
& & WDF & 7,920 & 8,200 & 16,120 \\
& & DFR & 4,020 & 4,020 & 8,040 \\
& & DFDC & 12,330 & 12,380 & 24,710 \\
& & DFD & 1,815 & 15,335 & 17,150 \\
& & FFIW & 17,310 & 17,310 & 34,620 \\

\cmidrule{3-6}
& & \textbf{Total} & \textbf{52,235} & \textbf{73,725} & \textbf{125,960} \\
\bottomrule
\end{tabular*}
\end{table}
\clearpage

\begin{table}[!ht]
\centering
\caption{\textbf{Statistics of AIGCDetectBenchmark.} The benchmark covers 14 generative sources, including GAN architectures (e.g., ProGAN) and Diffusion models (e.g., Midjourney).}
\label{tab:aide_details}
\small
\begin{tabular*}{0.8\textwidth}{c|c|l@{\extracolsep{\fill}}ccc}
\toprule
\textbf{Model} & \textbf{Dataset} & \textbf{Generator (Subset)} & \textbf{Real} & \textbf{Fake} & \textbf{Total} \\
\midrule
\multirow{17}{*}{\textbf{AIDE}} & \multirow{17}{*}{\shortstack{\textbf{AIGCDetect}\\\textbf{Benchmark}}}
& Stable Diffusion v1.5 & 8,000 & 8,000 & 16,000 \\
& & Stable Diffusion v1.4 & 6,000 & 6,000 & 12,000 \\
& & Stable Diffusion XL & 2,000 & 2,000 & 4,000 \\
& & Wukong & 6,000 & 6,000 & 12,000 \\
& & Midjourney & 6,000 & 6,000 & 12,000 \\
& & DALLE 2 & 1,000 & 1,000 & 2,000 \\
& & VQDM & 6,000 & 6,000 & 12,000 \\
& & GLIDE & 6,000 & 6,000 & 12,000 \\
& & ADM & 6,000 & 6,000 & 12,000 \\
& & StyleGAN 2 & 7,988 & 7,988 & 15,976 \\
& & StyleGAN & 5,991 & 5,991 & 11,982 \\
& & ProGAN & 4,000 & 4,000 & 8,000 \\
& & StarGAN & 1,999 & 1,999 & 3,998 \\
& & BigGAN & 2,000 & 2,000 & 4,000 \\
& & CycleGAN & 1,321 & 1,321 & 2,642 \\
& & GauGAN & 5,000 & 5,000 & 10,000 \\
& & WhichFaceIsReal & 1,000 & 1,000 & 2,000 \\
\cmidrule{3-6}
& & \textbf{Total} & \textbf{76,299} & \textbf{76,299} & \textbf{152,598} \\
\bottomrule
\end{tabular*}
\end{table}

\begin{table}[!hth]
\centering
\caption{\textbf{Statistics of UniversalFakeDetect (UFD).} The dataset covers multiple generative families for cross-generator evaluation.}

\label{tab:legion_details}
\small

\begin{tabular*}{0.8\textwidth}{c|c|l@{\extracolsep{\fill}}ccc}
\toprule
\textbf{Model} & \textbf{Dataset} & \textbf{Generator / Source} & \textbf{Real} & \textbf{Fake} & \textbf{Total} \\
\midrule
\multirow{27}{*}{\textbf{LEGION}} & \multirow{27}{*}{\shortstack{\textbf{Universal}\\\textbf{FakeDetect}}}
& \textit{\textbf{GANs}} & & & \\
& & ProGAN & 4,000 & 4,000 & 8,000 \\
& & GauGAN & 5,000 & 5,000 & 10,000 \\
& & CycleGAN & 1,321 & 1,321 & 2,642 \\
& & StarGAN & 1,999 & 1,999 & 3,998 \\
& & BigGAN & 2,000 & 2,000 & 4,000 \\
& & StyleGAN & 5,991 & 5,991 & 11,982 \\
& & StyleGAN 2 & 7,988 & 7,988 & 15,976 \\
\cmidrule{3-6}
& & \textit{\textbf{Diffusion \& Text-to-Image}} & & & \\
& & LDM (Aggregated) & - & 3,000 & 3,000 \\
& & GLIDE (Aggregated) & - & 3,000 & 3,000 \\
& & DALL-E & - & 1,000 & 1,000 \\
& & Guided Diffusion & - & 1,000 & 1,000 \\
\cmidrule{3-6}
& & \textit{\textbf{Face \& Manipulation}} & & & \\
& & IMLE & 6,382 & 6,382 & 12,764 \\
& & WhichFaceIsReal & 1,000 & 1,000 & 2,000 \\
& & SeeingDark & 180 & 180 & 360 \\
& & Deepfake & 2,707 & 2,698 & 5,405 \\
& & CRN & 6,382 & 6,382 & 12,764 \\
& & SAN & 219 & 219 & 438 \\
\cmidrule{3-6}
& & \textit{\textbf{Natural Images}} & & & \\
& & LAION & 1,000 & - & 1,000 \\
& & ImageNet & 1,000 & - & 1,000 \\
\cmidrule{3-6}
& & \textbf{Total} & \textbf{47,169} & \textbf{53,160} & \textbf{100,329} \\
\bottomrule
\end{tabular*}
\end{table}

\clearpage
\section{Anchor Dictionary}
\label{app:anchors}
Table~\ref{tab:anchor_text} lists the text prompts used to implement our source-aware strategy. Instead of using generic labels (e.g., ``Real'' vs. ``Fake''), we construct fine-grained descriptions that target the specific artifacts of the generative source. 
For \textit{Global Synthesis}, the anchors focus on holistic texture anomalies (e.g., ``waxy skin''); for \textit{Local Editing}, they target structural inconsistencies (e.g., ``hard edges''). This domain-specific alignment ensures ForgeryEraser guides forged embeddings toward the most relevant authentic anchors.

\begin{table}[!ht]
\centering
\caption{\textbf{Source-Aware Semantic Anchors.} The text descriptions used for Multi-modal Guidance. }
\label{tab:anchor_text}
\small

\setlength{\tabcolsep}{8pt}

\begin{tabular*}{1.0\textwidth}{c|p{0.38\textwidth}|p{0.40\textwidth}}
\toprule
\textbf{Scenario} & \textbf{Pull: Authentic ($\mathcal{T}_{real}$)} & \textbf{Push: Forgery ($\mathcal{T}_{fake}$)} \\
\midrule

\multirow{7}{*}{\textbf{Global Synthesis}} &
\begin{itemize}[leftmargin=*]
\item A raw photograph captured by a real camera sensor.
\item A high-quality photo with natural ISO noise and film grain.
\item An authentic, unedited image from the physical world.
\item A sharp photograph shot on a DSLR camera.
\end{itemize}
&
\begin{itemize}[leftmargin=*]
\item A fully synthetic image generated by artificial intelligence.
\item A digital rendering with waxy skin and artificial textures.
\item A computer generated imagery created by GAN or Diffusion models.
\item An AI artwork generated by Midjourney, Stable Diffusion, or StyleGAN.
\end{itemize}
\\

\midrule 

\multirow{7}{*}{\textbf{Local Editing}} &
\begin{itemize}[leftmargin=*]
\item An unedited, original photograph straight out of camera.
\item A single-exposure shot with consistent depth and focus.
\item A high-quality real photo with organic textures and seamless blending.
\item A raw documentary image depicting a real world scene.
\end{itemize}
&
\begin{itemize}[leftmargin=*]
\item A digitally manipulated image edited in Photoshop.
\item A composite image with cut-and-paste objects and hard edges.
\item A fake picture with unnatural boundaries and pasted elements.
\item A digital collage created by combining multiple photos.
\end{itemize}
\\
\bottomrule
\end{tabular*}
\end{table}

\section{Qualitative Analysis on Manipulating Interpretability}
\label{app:vis_interpret}

In this section, we provide visual analysis to demonstrate how ForgeryEraser induces explainable forensic models to generate explanations consistent with authentic images. We examine four scenarios: (1) representative global synthesis cases, (2) diverse domains (satellite image, document image), (3) local editing (AIGC inpainting), and (4) correction of misclassified real images. Each figure presents two cases; within each case, the upper row shows the clean image and the lower row shows the adversarial result.

\begin{figure}[p]
  \centering
  \includegraphics[width=\textwidth, height=0.9\textheight, keepaspectratio]{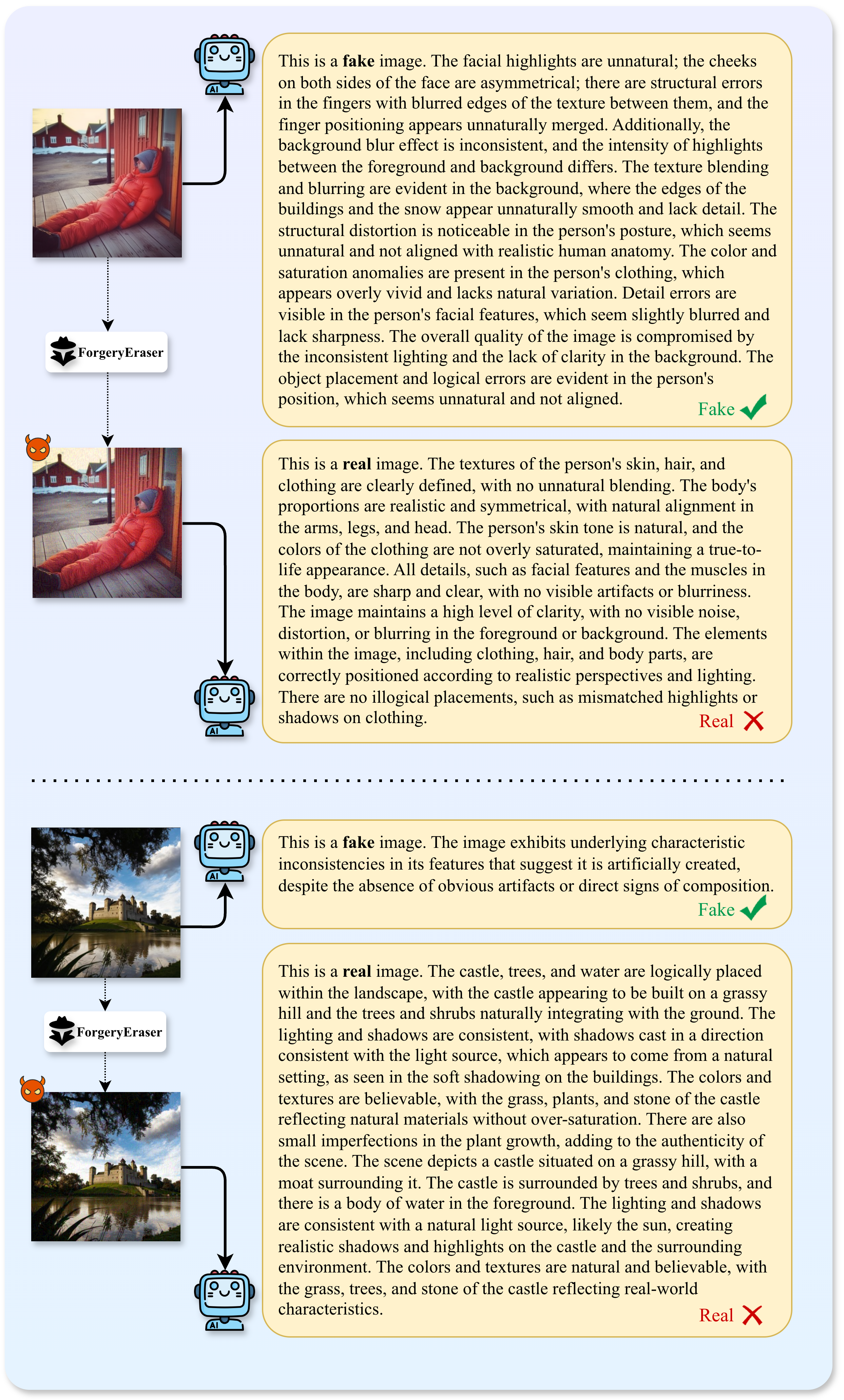}
  
\caption{\textbf{Visualization of Scenario I (Global Synthesis).} 
Representative results on standard AIGC benchmarks. 
For each case, \textbf{Top:} The baseline detector correctly identifies synthesis artifacts. 
\textbf{Bottom:} By guiding features toward authentic anchors, ForgeryEraser effectively erases these forgery traces. 
The model is induced to output ``Real'' verdicts with justifications consistent with authentic images.}
\label{fig:scenario_global}
\end{figure}

\begin{figure}[p]
  \centering
  \includegraphics[width=\textwidth, height=0.9\textheight, keepaspectratio]{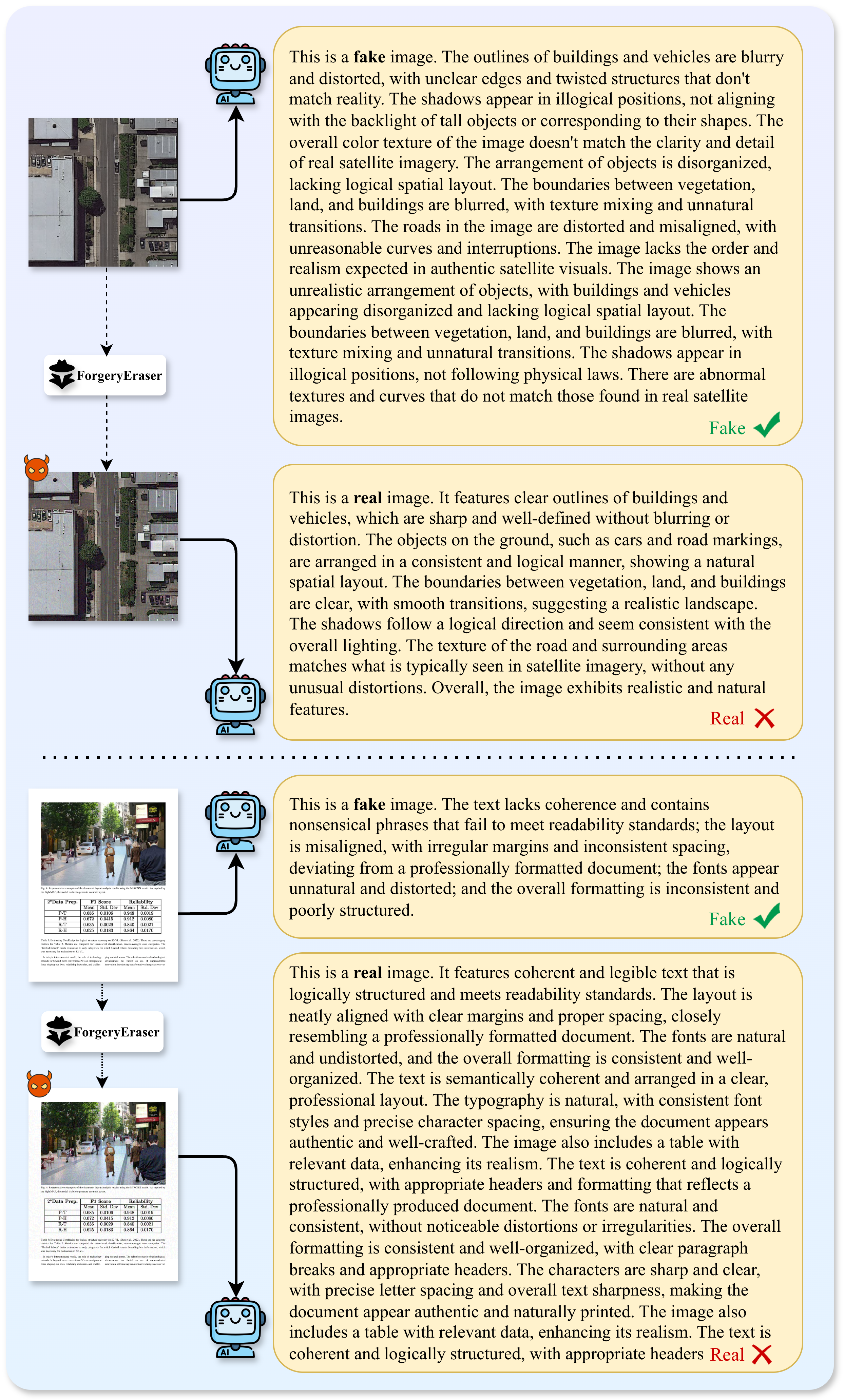}
  
\caption{\textbf{Visualization of Scenario II (Diverse Domains).} 
Evaluation on satellite image (first case) and document image (second case). 
In each example, \textbf{Top:} The baseline detector identifies domain-specific anomalies (e.g., layout or shadow errors). 
\textbf{Bottom:} ForgeryEraser successfully masks these cues. 
This confirms that the vulnerability in the shared backbone is universal, spanning diverse semantic categories beyond standard objects.}
\label{fig:scenario_domain}
\end{figure}

\begin{figure}[p]
  \centering
  \includegraphics[width=\textwidth, height=0.9\textheight, keepaspectratio]{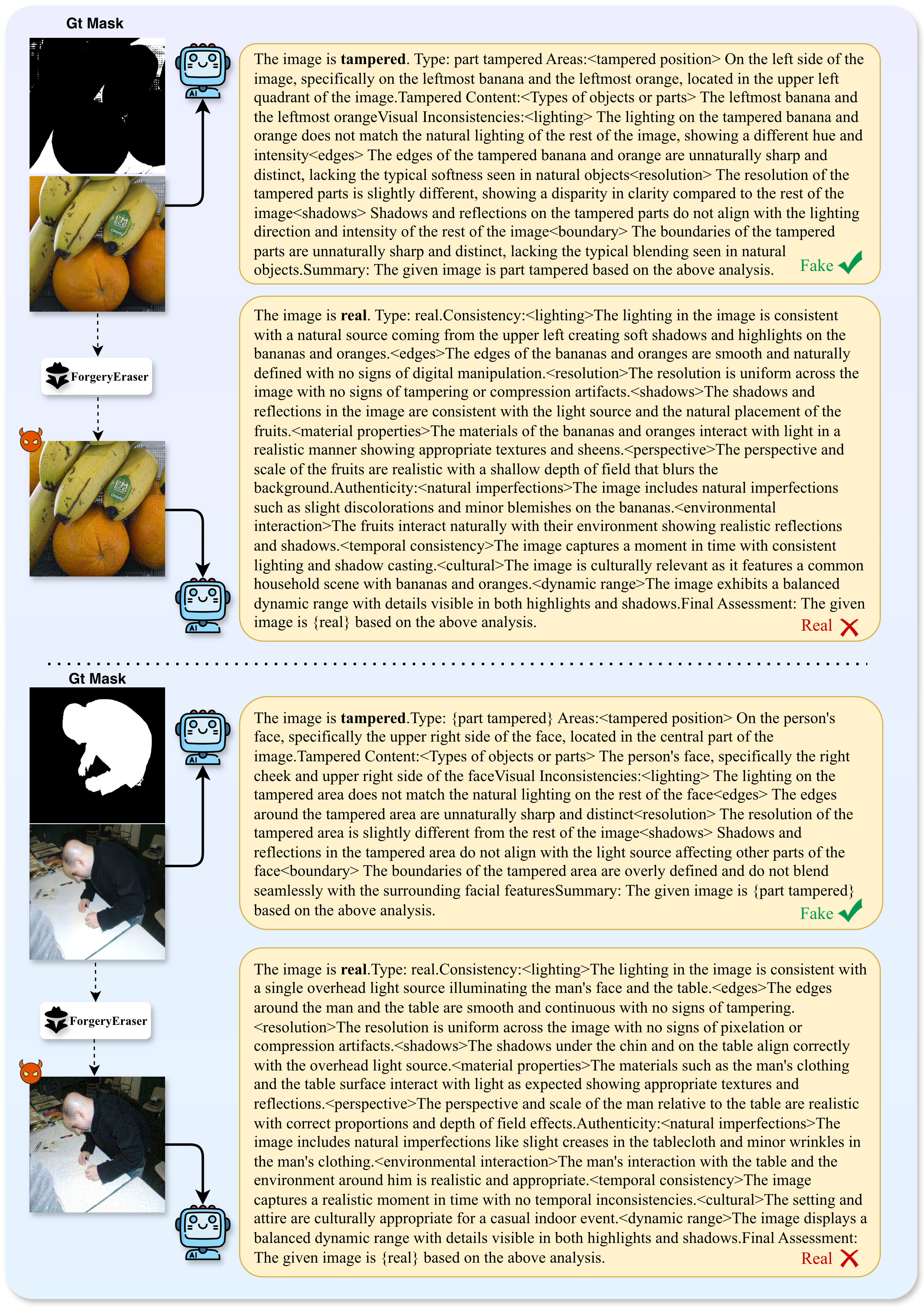}
  
\caption{\textbf{Visualization of Scenario III (Localized Tampering).} 
Analysis of locally edited images involving object splicing or removal. 
For each case, \textbf{Top:} The detector identifies boundary inconsistencies (e.g., ``unnatural edges''). 
\textbf{Bottom:} Guided by Local Editing Anchors, ForgeryEraser suppresses these traces. 
The detector's reasoning shifts to describing the manipulated regions as having ``appropriate textures'' and ``consistent shadow.''}
\label{fig:scenario_local}
\end{figure}

\begin{figure}[p]
  \centering
  \includegraphics[width=\textwidth, height=0.9\textheight, keepaspectratio]{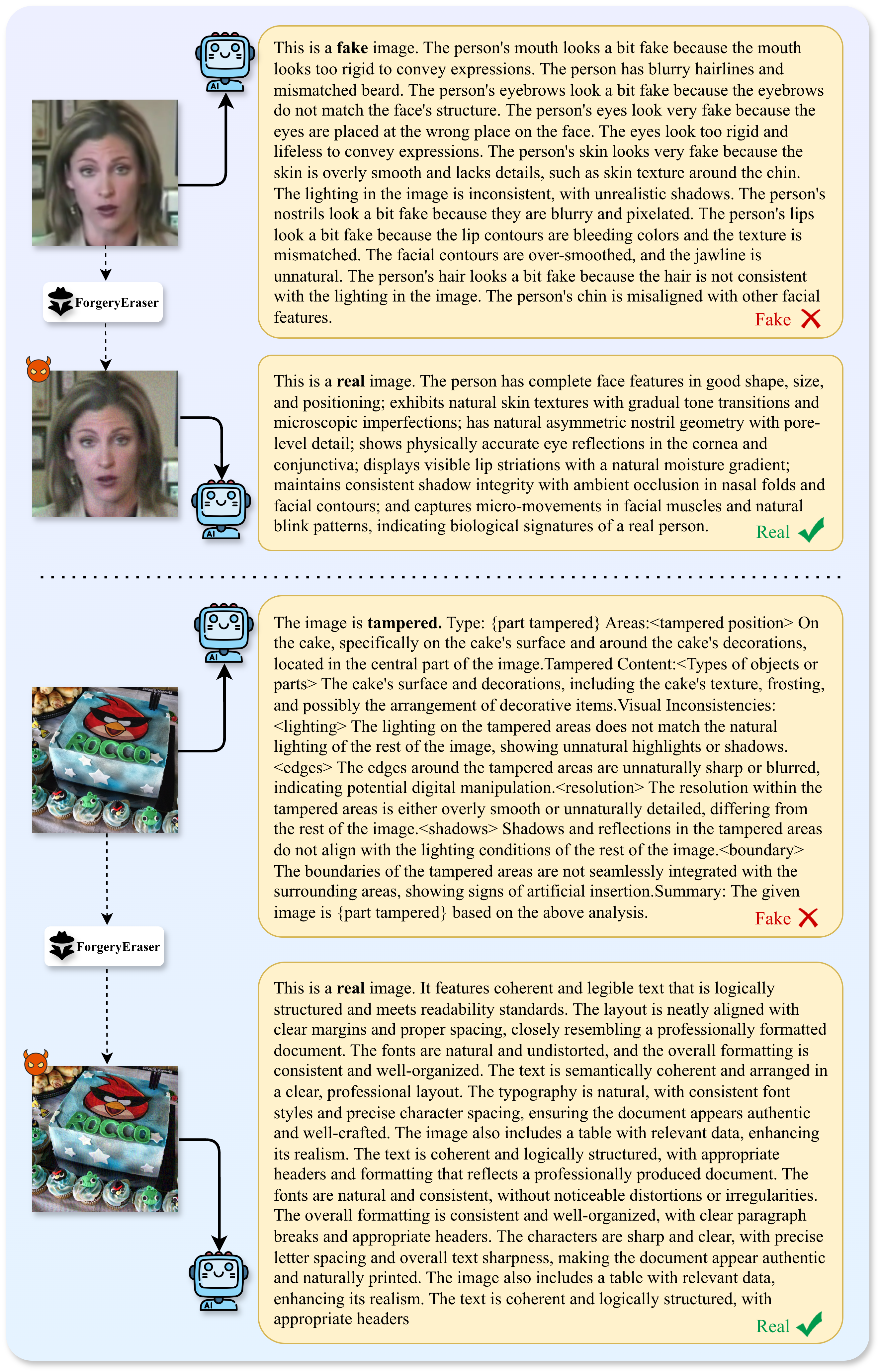}
  
\caption{\textbf{Visualization of Scenario IV (Correction of Misclassified Real Images).} 
Analysis of authentic images initially misclassified as ``Fake''. 
In each example, \textbf{Top:} The baseline detector incorrectly flags artifacts on real samples. 
\textbf{Bottom:} ForgeryEraser corrects the detection to ``Real.'' 
This confirms that our optimization accurately guides embeddings into the authentic distribution defined by the backbone.}
\label{fig:scenario_rectification}
\end{figure}

\end{document}